\documentclass[pdflatex,sn-basic]{sn-jnl}% Basic Springer Nature Reference Style/Chemistry Reference Style
\jyear{2022}%

\theoremstyle{thmstyleone}%
\theoremstyle{thmstyletwo}%

\theoremstyle{thmstylethree}%

\usepackage{silence}
\usepackage[style=base]{caption}
\usepackage{subcaption}
\usepackage[shortlabels]{enumitem}
\begin{document}

\title{FedUKD: Federated UNet Model with Knowledge Distillation for Land Use Classification from Satellite and Street Views}

%%=============================================================%%
%% Prefix	-> \pfx{Dr}
%% GivenName	-> \fnm{Joergen W.}
%% Particle	-> \spfx{van der} -> surname prefix
%% FamilyName	-> \sur{Ploeg}
%% Suffix	-> \sfx{IV}
%% NatureName	-> \tanm{Poet Laureate} -> Title after name
%% Degrees	-> \dgr{MSc, PhD}
%% \author*[1,2]{\pfx{Dr} \fnm{Joergen W.} \spfx{van der} \sur{Ploeg} \sfx{IV} \tanm{Poet Laureate} 
%%                 \dgr{MSc, PhD}}\email{iauthor@gmail.com}
%%=============================================================%%

\author[1]{\fnm{Renuga} \sur{Kanagavelu}}
\author[2]{\fnm{Kinshuk} \sur{Dua}}
\author[2]{\fnm{Pratik} \sur{Garai}}
\author*[2]{\fnm{Susan} \sur{Elias}}\email{susan.elias@vit.ac.in}
\author[3]{\fnm{Neha} \sur{Thomas}}
\author[4]{\fnm{Simon} \sur{Elias}}
\author[1]{\fnm{Qingsong} \sur{Wei}}
\author[1]{\fnm{Goh Siow Mong} \sur{Rick}}
\author[1]{\fnm{Liu} \sur{Yong}}

\affil[1]{\orgdiv{Institute of High Performance Computing}, \orgname{A*STAR},  \orgaddress{\country{Singapore}}}
\affil*[2]{ \orgname{Vellore Institute of Technology}, \orgaddress{\city{Chennai}, \country{India}}}
\affil[3]{\orgdiv{College of Engineering}, \orgname{Anna University}, \orgaddress{\city{Chennai}, \country{India}}}
\affil[4]{\orgname{Measi Academy Of Architecture}, \orgaddress{\city{Chennai}, \country{India}}}

%%==================================%%
%% sample for unstructured abstract %%
%%==================================%%

\abstract{Federated Deep Learning frameworks can be used strategically to monitor Land Use locally and infer environmental impacts globally. Distributed data from across the world would be needed to build a global model for Land Use  classification. The need for a Federated approach in this application domain would be to avoid transfer of data from distributed locations and save network bandwidth to reduce communication cost. We use a Federated UNet model for Semantic Segmentation of satellite and street view images.  The novelty of the proposed architecture is the integration of Knowledge Distillation to reduce communication cost and response time. The accuracy obtained was above 95\% and we also brought in a significant model compression to over 17 times and 62 times for street View and satellite images respectively.  Our proposed framework has the potential to be a game-changer in real-time tracking of climate change across the planet.}


\maketitle

\section{Introduction}
 Monitoring Land Use and Land Cover (LULC) in real-time, for tracking environmental impact across the globe, is required for the sustainable development of our planet. The speed at which the problem is advancing, combined with an urgent need to address already forming ecological impacts of climate change, has spurred fast-moving research into the field. With the rapid rise of climate change as the world’s biggest environmental challenge, the worldwide predicament of how to efficiently combat it,  is the focus currently. Around 196 countries had entered the Paris climate accord in 2015, an international agreement that pledges to keep the global average temperature rise to below 1.5°C and to reduce greenhouse gas emissions. The challenge however, is to efficiently monitor compliance and bring in accountability on a global scale.  We propose the design of an AI driven model to monitor ground truth of Land Use,  in order  to estimate its impact on environment in real-time. 
\par There are two different ways in which LULC can be monitored:  (i) using aerial views from satellites or drones and (ii) using street views from CCTV cameras.  From an aerial view, we can observe the overall change in land cover and obtain environmental data across different types of land. The street view analysis,  provides city-wide insights on energy consumption, population density etc. The two views combined can provide  extensive analysis of land use, for detection and monitoring, from two different perspectives.  Automating the surveillance of LULC globally can therefore help to monitor climate change that can be inferred from Land Use.
\par Satellite views of countries across the world can vary dramatically in their characterisation. Even within a country there could be land categories that range from snow covered areas to deserts, agricultural areas and water bodies besides urban and non-urban residential and commercial areas. Similarly the objects that feature in street views can drastically change based on local culture, climate and   economic status of the region. The data will therefore be non-IID (Independent and Identically Distributed) having feature distribution skew, label distribution skew, quantity skew, etc. 
\par We propose a Federated Learning (FL) approach to train a Deep Leaning model for Semantic Segmentation. The motivation to use FL is to be able to deal with the non-IID nature of data from across the globe.  When models are built using publicly available images, data privacy or security is not the issue but increased communication cost due to high network bandwidth requirement will be a bottle-neck if data needs to be available in a centralised location for training.  In real-time when the model is created using distributed data from various countries, communication cost needs to be minimal in order to claim that the model is efficient. In order to mitigate all these issues, in this paper a novel Federated UNet with Knowledge Distillation (FedUKD) is presented. 
\section{Related work in Land Use Classification}
Several research work in the past have focused on automating the task of Land Use and Land Cover (LULC) detection for real-time monitoring and surveillance. Machine learning (ML) and Deep Learning (DL)  approaches have been used extensively in this domain and insightful research findings have been reported. Amongst the various state-of-the-art image processing techniques, \textit{semantic segmentation} has been the most relevant approach for Land Use and Land Cover (LULC) change detection. The spatial resolutions of satellite images made available for research has been improving, adding several new dimensions to the field as smaller urban objects are now visible. The research approaches used earlier were, spectral image classification, pixel-based image analysis (PBIA) and object-based image analysis (OBIA), but currently there seems to be a paradigm shift towards pixel-level semantic segmentation. A recent research publication \cite{SV006}, has presented a comprehensive review of the advancements in deep learning-based semantic segmentation for urban Land Use and Land Cover (LULC) detection. The main task of semantic segmentation is to classify each pixel of an image, using features that are derived from the ground truth indicated using masks in annotated images. DL frameworks  use the masks as labels for automated selection of features that enable classification of LULC regardless of seasonal, temporal and spatial changes \cite{UB001, UB004, UB010}.

The survey presented in \cite{UB005} has documented details of  DL frameworks that are the foundation for many modern image segmentation DL architectures.  One popular framework is the encoder-decoder based models,  with the UNet being of significant importance. It consists of two functional blocks; a contracting or down-sampling block which extracts features using 3x3 convolutions. The features are then copied to the expanding or up-sampling block which reduces feature maps and increases dimensions, so as to not lose pattern information. From the feature maps a segmentation map is then generated. It is worth noting that UNets make use of data augmentation, which increases the amount of labeled samples used in training, as most often the datasets used may not contain enough annotated data required for training the model from scratch. It is thus used to learn from very few labeled images. In the work presented in this paper we make use of UNet based DL architecture for semantic segmentation. Several existing research work have used image segmentation for Land Use and Land Cover (LULC) classification towards the goal of monitoring changes in urban Land Use through satellite images and street view images, involving several different Land Use categories. The following subsections summarize these work.  

\subsection{LULC detection from Satellite Images}
In several existing works \cite{UB008, UB013, UB018} the use of DL techniques in LULC classification, emphasizing on multispectral and hyperspectral images are illustrated. These types of images contain spatial and spectral information, from which material characteristics and differences in the Land Use categories are retrieved for classification. The data sets used for this were both extensive with ample labelled images as well as datasets containing images that were to be labelled using semi-supervised techniques, as only a relatively  few real-world data samples can be  provided along  with corresponding  ground truth information. It was found that aerial remote sensing images had a higher spatial quality, so they were preferred albeit with the limitation of low temporal resolution. The challenges were in obtaining well-annotated data-sets as well as  remote sensing technologies that worked with the deep learning frameworks. Additionally, for any deep learning model to learn and leverage spatial patterns, combining the mapped images with the ground truth is necessary. This mapping process can been done in several ways \cite{UB016} and most commonly using a three-dimensional grid that conserves the spatial relationships between the different landmarks and locations in the imagery. These two sets of images, the mapped imagery as well as the ground truth, are used as training samples and for automatic feature extraction by the DL model \cite{UB014}.  An interesting work that monitors landscape changes in coastal, agricultural as well as urban areas is presented in \cite{UB018}.  Satellite images   were segmented and classified  into categories such as wetland, water, agriculture, and urban/suburban regions. Using the OpenStreetMap (OSM) database, a few existing research work \cite{UB006, UB017} have demonstrated the classification of Land Use categories such as water, road, agricultural, with seasonal and climate-related changes. Time-series analysis is also relevant to aerial/satellite image analysis, with algorithms such as Dynamic Time Warping (DTM) filling in temporal gaps in the remote sensing time-series data \cite{UB009} and helping to overcome the limitation of irregularly distributed training samples. Deep Learning frameworks for classifying crops with cloudy and non-cloudy images using the Sentinel-2 time series dateset have been presented\cite{UB011, UB012}. Other works \cite{UB015, UB002} have focused on changes in water bodies to monitor and analyze the urban aquatic ecosystem. This was done by using  remote-sensing technologies such as Normalized Difference Water Index (NDWI) and Modified Normalized Difference Water Index (MNDWI) methods for classification of water bodies and non-water bodies. 

\subsection{Street View Image Analysis}
Semantic segmentation on street view images has been an interesting area of research.  The most popular models for semantic segmentation are FCN (Fully Convolutional Network), UNet and DeepLabV3+. In \cite{SV001} a comparative analysis of the performance on these models have been done and presented for 11 classes of street objects such as sky, building, road, tree, car, pavement, pedestrian etc. The dataset used was CamVid (2007) one of the first to provide a labelled data set for semantic segmentation. The results showed that each of the above three models performed better for some classes in comparison to the others. The work presented in \cite{SV002} uses the Tencent Map dataset, to analyse the percentage of green and blue cover available in street view images. Semantic segmentation of the images were done to detect the percentage of green and blue cover available in street views. This information was correlated with the mental health data for the corresponding regions. The goal was to  analyse the impact of these outdoor features on geriatric depression. 

 In \cite{SV003} Google Street View data was used to  develop an automated street sidewalk detection model. The images were translated into  graph-based segments and  machine learning approaches were used to obtain good accuracy. The goal of the work presented in \cite{SV004} was to design a model for real-time semantic segmentation of street views required for autonomous applications. The model called DSANet was introduced and found to have faster response during real time inference of semantic segmentation on street scenes taken from CamVid and Cityscape data sets \cite{SV011}.  Analysing  the model performance on the Mapillar Vista dataset \cite{SV005}  was considered challenging as the images were harder to segment compared to previously released datasets for semantic segmentation. A variant of the FCN called Dilated Convolution Network was designed to meet this challenge and it was found to have good accuracy for segments occupying large areas. Another challenging set of street view images containing changes in scenes due to seasonal and lighting variations was released in \cite{SV007}  and a model for pixel-wise change detection occurring due to season, lighting and weather called Deep Deconvolutional Network was also presented. 

 In \cite{SV008} a global-and-local network architecture (GLNet) has been proposed with the objective of incorporating  context information from local context and  spatial information from a global context for better performance. This was analysed using the Cityscape dataset for street view and the improvements over the existing models including PSPNet, ICNet are presented. A comprehensive review of the various data sets that can be used and research directions in the field are also well documented \cite{SV009}. However there has been no attempt so far in designing a Federated Deep Learning architecture for semantic segmentation of street view images. This is the research gap that has been addressed in our paper. 
 
\section{Material and methods}
%\subsection{Federated Learning for Land Use Classification}
 \begin{figure}
  \includegraphics[width=\linewidth]{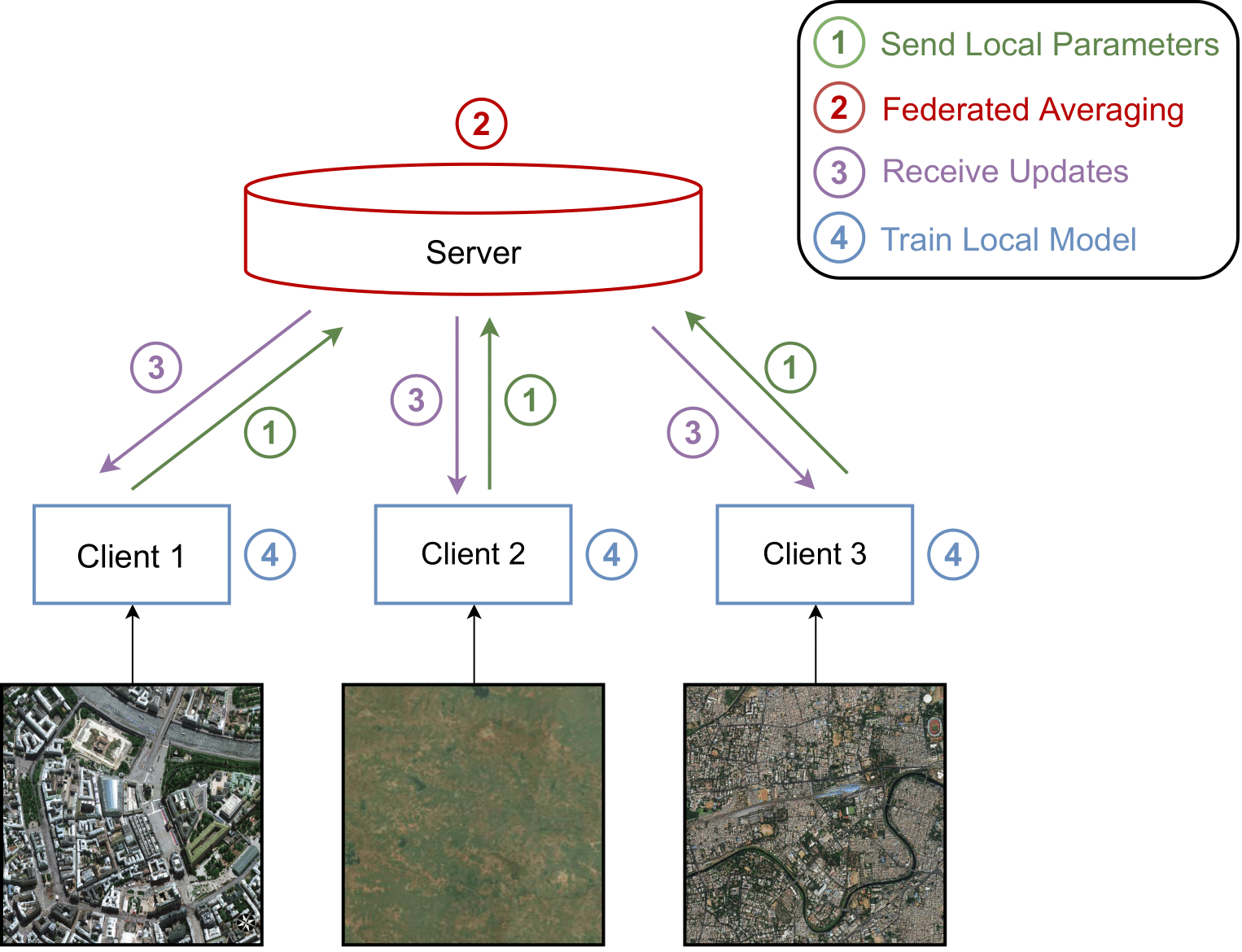}
  \caption{FL Model for Semantic Segmentation of Satellite Images}
   \label{fig:Fed}
  \end{figure}
Federated Learning (FL)   is an emerging field with many interesting open research problems \cite{FL001}. It is expected to provide \textit{Responsible AI} solutions by integrating data privacy and security measures. 
But despite not sharing data in distributed computing systems, there are  several privacy threat models associated with the different stages of ML/DL. Data publishing stage is prone to \textit{Attribute Inference} attacks, model training  stage is vulnerable to \textit{Reconstruction} attacks while model inference stage can be affected by \textit{Reconstruction attacks}, \textit{Model Inversion} attacks and \textit{Membership Inference} attacks. Privacy preserving ML/DL use defence techniques such as  \textit{Homomorphic Encryption}, \textit{Secure Multi-Party Computation} and \textit{Differential Privacy}.  For the application proposed in this paper,  the goal  is to adopt a Federated Learning approach in developing a shared deep learning model for semantic analysis of satellite and street view images. Federated Averaging Algorithm (FedAvg) was first employed for distributed machine learning over horizontally partitioned dataset \cite{FL002}. In Federated Averaging, each party uploads clear-text gradient to a coordinator independently, then the coordinator computes the average of the gradients and updates the model. Finally, the coordinator sends the clear-text updated model back to each party. Federated Averaging (FedAvg) algorithm works well for certain non-convex objective functions under Independent and Identical (IID) setting but may not work in non-IID datasets \cite{FL003}. 
\subsection{Federated UNet for Semantic Segmentation}
The motivation to create Federated UNet model for Land Use classification was derived from the work presented in a related work titled \textit{UNet for Semantic Segmentation on Unbalanced Aerial Imagery}(https://github.com/amirhosseinh77/UNet-AerialSegmentation).
The related repository presents a dataset created from the UAE region, with labels intended for semantic segmentation of Land Use.  Land Use can vary from country to country based on the geographical terrain. Some countries may have more of desserts, forests, snow covered land, agricultural land, water bodies etc besides the regular residential and non-urban areas. For some countries the same Land Use can look very different in the aerial view, for instance residential areas could range from chaotic looking unplanned areas to well planned urban areas in some countries. These real-world variations make Land Use classification very region specific. In order to have global models to monitor climate change across the world we need to have a Federated Learning approach in training the models toward automation in Land Use detection. We first implemented a Federated UNet model using the datasets and distributed scenarios described in the following section. The UNet used was a big model with about 17 million parameters and hence communication overhead was high. The accuracy was good and the model was found to be effective for a Federate Learning framework. Thus the Federated UNet model was  used for LULC classification while a  further enhancement  through model optimization was also achieved and the details are presented in the following subsection. 

\subsection{FedUKD: Federated Unet with Knowledge Distillation}
\par We have successfully implemented a Federated UNet based architecture for semantic segmentation of satellite images and street view images. We have used a 3 client scenario to demonstrate the model performance. But in reality the design of a global model for LULC detection  will involve contributions by thousands of local clients models from participating  countries leading to high communication overheads. In order to optimize communication cost,  we have integrated  a Teacher-Student Knowledge Distillation approach presented in the FedKD model \cite{FL004}. The reduction in communication cost is achieved by sharing the smaller student model to build the global model instead of sharing large models with the server. The larger teacher model is used locally for reciprocal learning between the teacher and student models. The teacher model is updated and maintained  locally at each client while the student model updates are all aggregated by the central server and global updates are distributed to clients to update their student models. This process continues till the student model converges. 
 \begin{figure}
  \includegraphics[width=\linewidth]{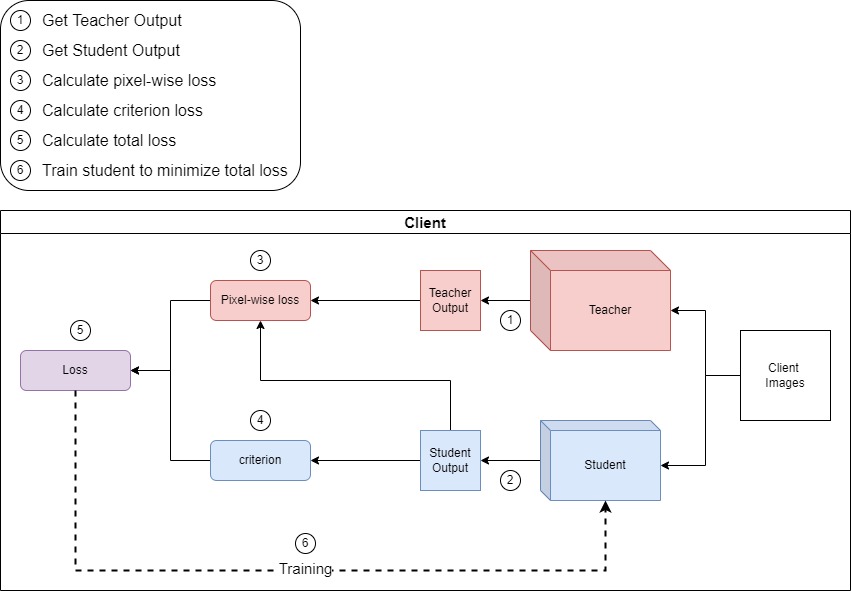}
  \caption{Teacher-Student Federated Learning Model Architecture}
   \label{fig:kd}
  \end{figure}
\par In Figure \ref{fig:kd} the Teacher-Student Knowledge Distillation workflow is illustrated. This is the interaction that happens at every client location. For the Teacher we use an UNet having around 17 million parameters. We implemented the Knowledge Distillation framework for a larger  CitytScape dataset.  For the Student model we used an approach to downsize the UNet presented in an existing work \cite{FL005, FL006}. We proposed a much smaller FCN network that was derived by taking the architecture of the original UNet and reducing the number of layers and parameters in the structure. The kernel size of the convolutional layer, was kept the same, i.e., 3x3. Our proposed model has 2 downsample and upsample layers each (vs 4 of each in the original UNet). The number of filters in these convolution blocks were reduced to (16, 32, 32, 16) vs (64, 128, 256, 512, 512, 256, 128, 64) of the UNet model. This reduction in the number of layers and the filters in these layers gave us a much smaller model with ~1,18,000 parameters. The performance of the models and related discussions are presented in the following subsections. Algorithm 1, describes the model building process and details of the communication rounds. The procedure \textit{Server} describes the global update rounds and aggregation process and procedure \textit{UKDLearning} explains about the Teacher-Student Knowledge Distillation process. The Algorithm 1 helps to understand the overall workflow of the proposed  FL framework for designing a UNet model through Knowledge Distillation to save communication cost. 

\begin{algorithm}
\small
\caption{Federated UNet with Knowledge Distillation}
\hspace*{\algorithmicindent} \textbf{Input:} pretrained Teacher T \\
 \hspace*{\algorithmicindent}\hspace*{\algorithmicindent}\hspace*{\algorithmicindent} number of Clients C \\
  \hspace*{\algorithmicindent}\hspace*{\algorithmicindent}\hspace*{\algorithmicindent} number of communication rounds R \\
 \hspace*{\algorithmicindent}\hspace*{\algorithmicindent}\hspace*{\algorithmicindent}  number of local training iteration i \\
\hspace*{\algorithmicindent} \textbf{Output:} Globally trained model weights  $w^{r}_{f}$ and $w^{r}_{g}$ 

\begin{algorithmic}[1]
\Procedure{Server}{}    \\
\hspace*{\algorithmicindent}Initialize $w^{0}_{f}$ and $w^{0}_{g}$
\hspace*{\algorithmicindent}\For{i = 0, 1, ... R{-}1  }
\hspace*{\algorithmicindent}\hspace*{\algorithmicindent}\For{i = 0, 1, ... C  in parallel}\\
 \hspace*{\algorithmicindent}\hspace*{\algorithmicindent}\hspace*{\algorithmicindent}$(w^{r}_{f})_{i} \leftarrow UKDLearning(i, T (w^{r}_{f})_{i})$)
\EndFor\\
\hspace*{\algorithmicindent}\hspace*{\algorithmicindent}$(w^{r+1}_{g}), (w^{r+1}_{f})\leftarrow FedAvg(w^{r}_{f})$
\hspace*{\algorithmicindent}\hspace*{\algorithmicindent} \For{i = 0, 1, ... C  in parallel}\\
\hspace*{\algorithmicindent}\hspace*{\algorithmicindent}\hspace*{\algorithmicindent}send($(w^{r}_{f})_{i}$) to i\\
\EndFor
\EndFor
\EndProcedure
\Procedure{UKDLearning}{(i,t,$(w^{r}_{f})_{i}$):}\\
Update student based on $(w^{r}_{f})_{i}$
\hspace*{\algorithmicindent}\hspace*{\algorithmicindent}\For{n = 0, 1, ... E}
\hspace*{\algorithmicindent}\hspace*{\algorithmicindent}\hspace*{\algorithmicindent}\For{d in dataset}\\
\hspace*{\algorithmicindent}\hspace*{\algorithmicindent}\hspace*{\algorithmicindent}\hspace*{\algorithmicindent}  $S \leftarrow student(d)$\\
\hspace*{\algorithmicindent}\hspace*{\algorithmicindent}\hspace*{\algorithmicindent}\hspace*{\algorithmicindent}  $t \leftarrow T(d)$\\
\hspace*{\algorithmicindent}\hspace*{\algorithmicindent}\hspace*{\algorithmicindent}\hspace*{\algorithmicindent} $L_{P} \leftarrow pixelwiseloss (s,t)$\\
\hspace*{\algorithmicindent}\hspace*{\algorithmicindent}\hspace*{\algorithmicindent}\hspace*{\algorithmicindent} $L_{C}\leftarrow criterion(S,gt)$\\
\hspace*{\algorithmicindent}\hspace*{\algorithmicindent}\hspace*{\algorithmicindent}\hspace*{\algorithmicindent} $L_{t} \leftarrow \alpha L_{P} - (\alpha - 1)L_{C}$\\
\hspace*{\algorithmicindent}\hspace*{\algorithmicindent}Train student to minimise $L_{t}$
\EndFor
\EndFor
\EndProcedure
\end{algorithmic}
\end{algorithm}
\normalsize
\section{Experimental}
For the research presented in this paper we use separate datasets to demonstrate Federated Learning for aerial views and street views. For the aerial views we created our own dataset and for street view we used the Cityscape dataset. More details on the datasets are presented in the following subsections:
\subsection{Land Use Classification from Satellite Images}
\subsubsection{Proposed Chennai Land Use (CLU) Dataset}
\begin{figure}
  \includegraphics[width=\linewidth]{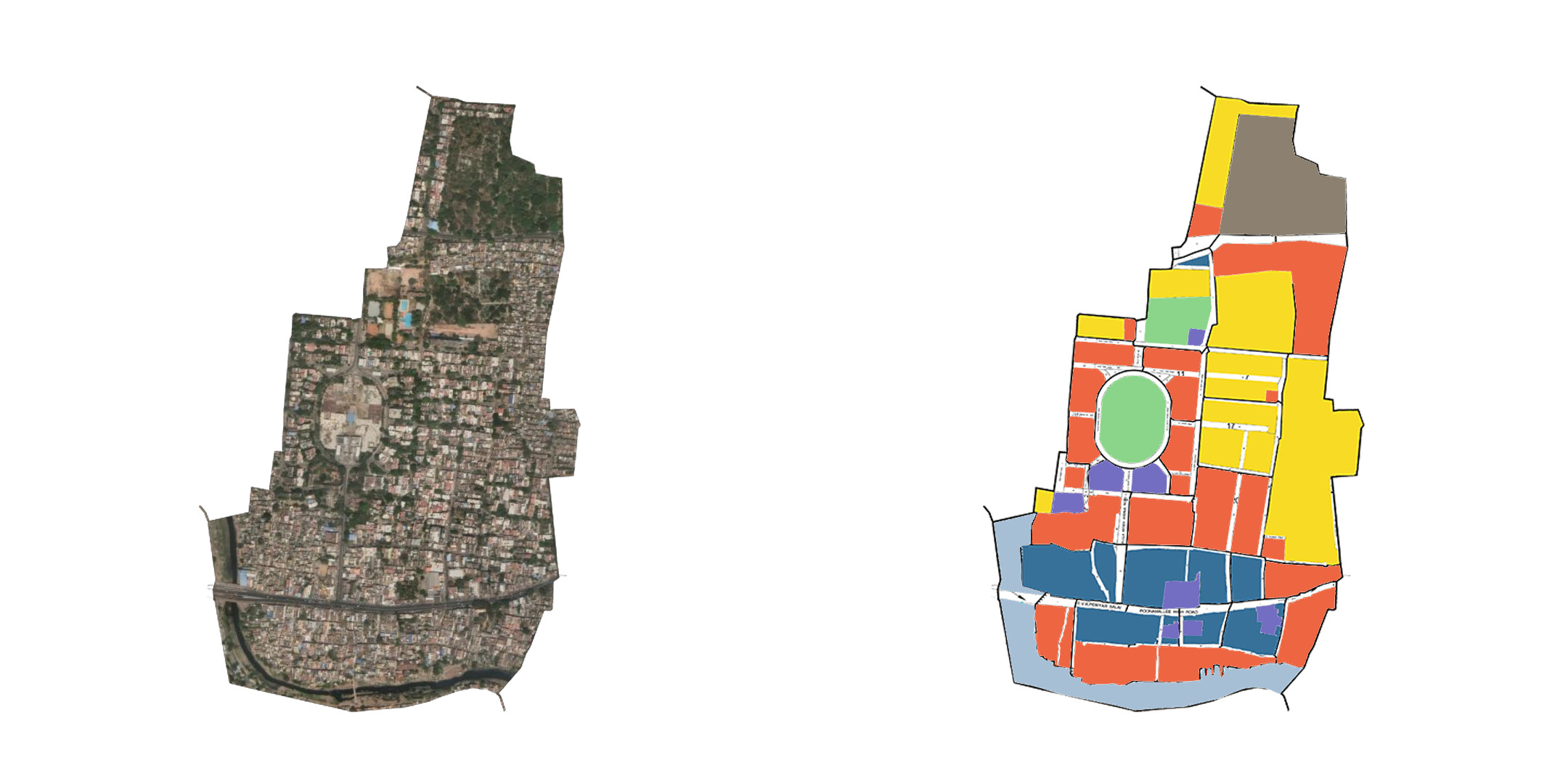}
  \caption{Satellite image and corresponding label - Chennai Land Use dataset}
   \label{fig:aminjikarai2}
  \end{figure}

Creating labels for semantic segmentation is a challenging task. In most cases particularly in medical imaging domain experts have limited time to spare to support  research activities. Moreover labelled segments could vary from person to person and hence post-processing techniques are employed to bring about a consensus amongst the labels provided by different experts. When  labels are difficult or impossible to obtain, approaches such as weakly supervised and semi supervised learning are used for training the models. Here we present a novel approach to creating Land Use image labels that can be generated in a consistent manner providing reliable and accurate labels for training. We use the satellite images of Chennai, a city in India for this demonstration. 
Chennai Metropolitan Development Authority (CMDA) has provided the city's Masterplan Land Use maps on their official website (http://cmdalayout.com/landusemaps/landusemaps.aspx). 
These maps provide the ground truth accurately. We used these maps region-wise and extracted the corresponding satellite images from the Google Earth portal. Since the original CMDA maps had legends that were not suitable for semantic segmentation, the colour coding had to be redone for each of the regions. For illustration, the satellite image and corresponding annotated image of a region called \textit{Aminjikarai} has been  presented in Figure \ref{fig:aminjikarai2}. The Land Use dataset we created contains labelled data for 70 such regions in Chennai. The dataset can be download and used for research purposes. (Link to download is available at the end of Section 5).
  
\subsubsection{CLU Data Set for Federated Learning}
In order to synthetically create a  label skewed  distributed scenario for the study, the Chennai Land Use (CLU) dataset was split into 3 datasets as follows:
\setlist{nolistsep}
\begin{itemize}[noitemsep]
\item Training - total 60  satellite images and corresponding ground truth images
    \begin{itemize}
        \item Dataset 1 - has 29 images. Images with legends \textit{water body, agricultural, urbanize} and \textit{Coastal Regulation Zone (CRZ)} can be found only in this dataset. 
        \item Dataset 2 - has 9 images. These images do not have the legends \textit{water body, agricultural, urbanize} and \textit{CRZ} but all of them have the legend \textit{non urban}. 
        \item Dataset 3 - has 22 images. These images do not have legends \textit{water body, agricultural, urbanize, CRZ} and \textit{non urban}
    \end{itemize}
\item Testing - total 10 images 
\end{itemize}
\subsection{Land Use Classification from Street View Images}
\subsubsection{CityScape Street View (CSP) Dataset}
In this research the dataset used to infer Land Use from street view images is  the CityScapes dataset \cite{SV011} created primarily to support research related to semantic understanding of urban street scenes.  It is a labelled dataset providing polygonal annotations for vehicle, people and objects in urban outdoors for about 30 classes. The dataset provides street views from 50 cities with diverse scenarios of seasons, weather, time of the day, as well as a range of objects with vibrant backgrounds and scene layouts. To create the data set needed for this research, we used samples from \textit{CityScape fine annotations} as segmentation masks for corresponding images from the \textit{left 8-bit} samples provided for  training.  
\subsubsection{CSP Data Set for Federated Learning}
The focus of the research presented in this paper is on the proposed Federated Learning (FL) framework for semantic analysis of street view images. 
We consider three different locations in a city to be the 3 clients in our Federated Learning framework. The images for training / inference  could be obtained from a localised CCTV footage  or Google Street View images captured in real-time.  In a smart city deployment this Federated Learning model could scale to thousands of client nodes representing all the streets of an entire city under surveillance. Sample images from \textit{CityScape} dataset are used for representation in this section. It has 30 labelled classes of  outdoor objects  and a subset  of the objects are present in each images. %In Figure \ref{fig:vehicle} the scene contains vehicles and a few other outdoor objects but there are no people in the scene. %
In order to create a label skewed scenario  for the study, we have considered  class \textit{vehicle} and class \textit{people}  and ensured that the dataset created for the 3 clients do not have an overlap of these labels. Some of the other classes may or may not be present based on the view. % In Figures \ref{fig:vehicle}, \ref{fig:people} and \ref{fig:road} the corresponding labelled images have been included for reference. % 
\begin{figure}
\centering
  \includegraphics[width=6.5cm]{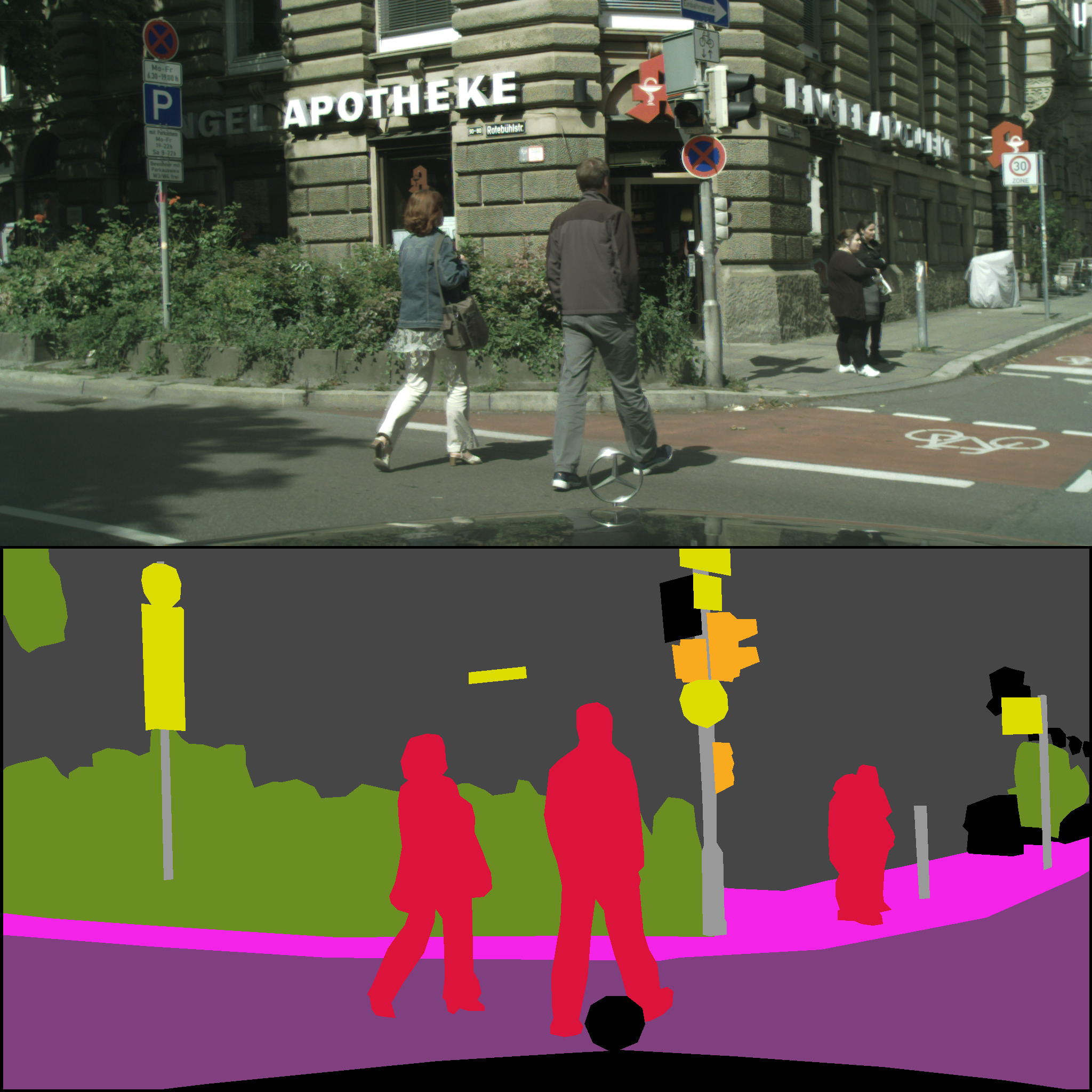}
   \caption{Street view with people but no vehicles}
 \label{fig:people1}
\end{figure}

In order to synthetically create a  label skewed  distributed scenario for the study, the \textit{CityScape} dataset was split into 3 datasets as follows:
\setlist{nolistsep}
\begin{itemize}[noitemsep]
\item Training - total 98 original images and corresponding ground truth images
    \begin{itemize}
       \item Dataset 1 - Vehicles with no people  - 65 images
        \item Dataset 2 - People with no vehicles  - 22 images
        \item Dataset 3 - No people and no vehicles  - 11 images
    \end{itemize}
\item Testing - total 20 images 
\end{itemize}
For analysing the model performance of our proposed FedUKD; the Federated UNet model with Knowledge Distillation, the CSP data was quantity skewed for the study. Here the training set considered had 2975 street view samples with corresponding masks and 500 images were available for validation. The motivation of the research is to develop a global model for street view understanding for challenging scenarios where different locations of a city present different subsets of objects to be trained or identified. Currently privacy-preserving approaches are  integral part of any distributed computing application and hence Federated Learning frameworks for real-time smart city surveillance are  being considered by governments globally.  

\section{Results and Discussions}
\begin{figure}
     \centering
     \begin{subfigure}[b]{0.20\textwidth}
      \includegraphics[width=\linewidth]{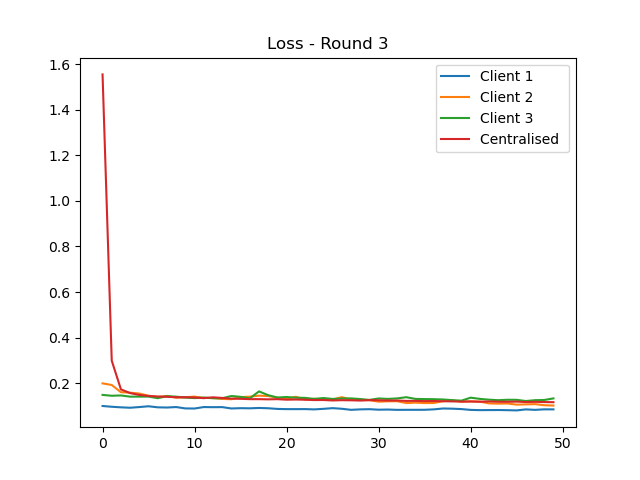}
      \caption{Chennai Land Use Dataset}
       \label{fig:CLU1}
      \end{subfigure}
         \begin{subfigure}[b]{0.20\textwidth}
      \includegraphics[width=\linewidth]{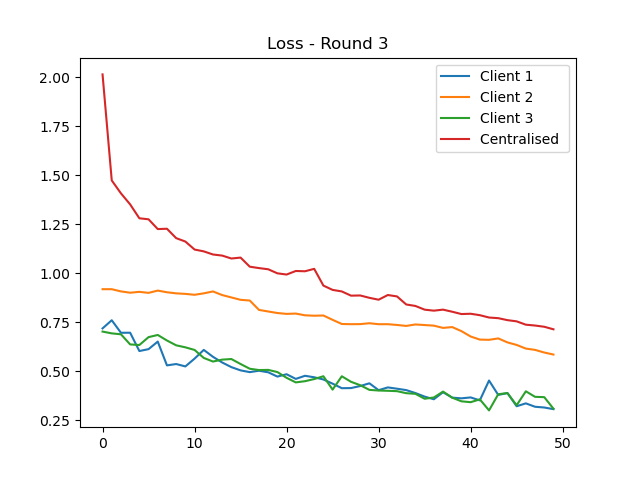}
      \caption{CityScape Dataset}
       \label{fig:CSP1}
      \end{subfigure}
     \caption{FedUnet - Model Performance}
\end{figure}

\begin{figure}
     \centering
      \subfloat[\centering CityScape Dataset]{{\includegraphics[width=0.4\linewidth]{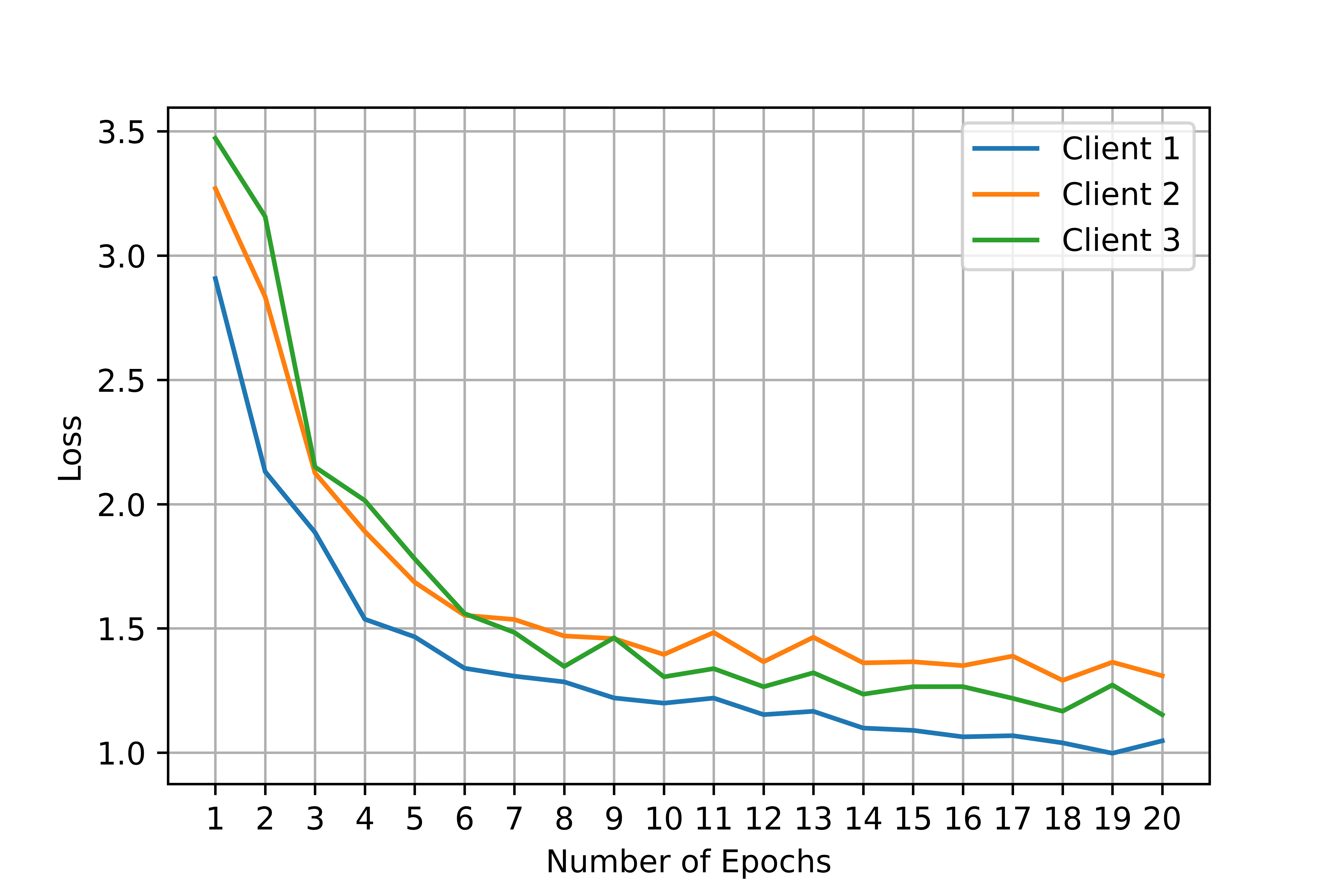}}      \label{fig:FedUKDCS}}%
       \qquad
      \subfloat[\centering Chennai Land Use Dataset Dataset]{{\includegraphics[width=0.4\linewidth]{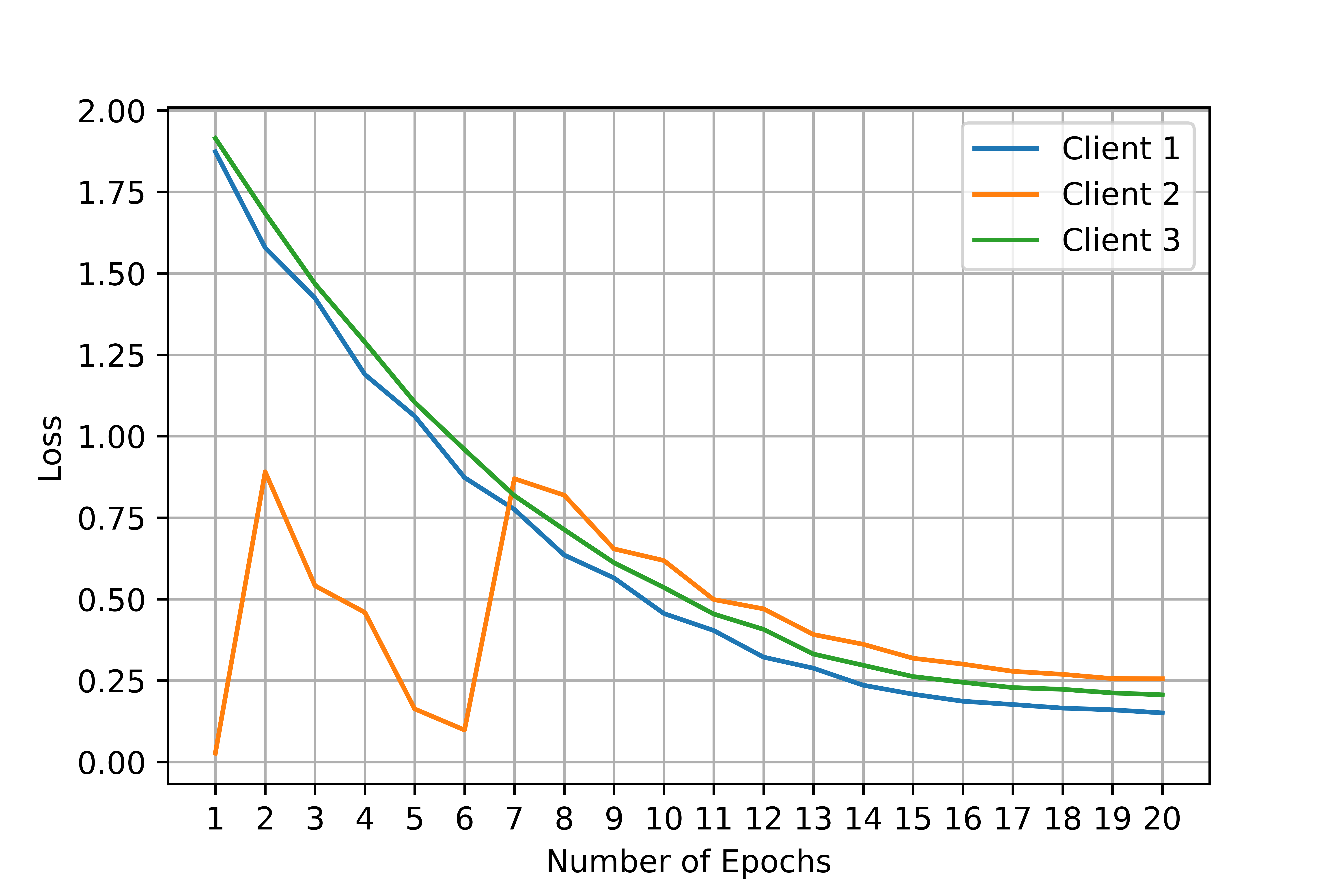}}     \label{fig:FedUKDCLU}}%
      \caption{FederatedUKD - Model Performance}
\end{figure}

Model performance analysis were carried out as follows: 
\setlist{nolistsep}
\begin{itemize}[noitemsep]
    \item Centralized UNet Performance on Chennai Land Use dataset
    \item Centralized UNet Performance on CityScape dataset
    \item Federated UNet Performance on label skewed  Chennai Land Use dataset
    \item Federated UNet Performance on label skewed CityScape dataset
    \item FedUKD Model Performance on  quantity skewed  Cityscape dataset
    \item FedUKD Model Performance on  label skewed  Cityscape dataset
\end{itemize}
The Chennai Land Use (CLU) dataset  comprising of 60 images were divided among 3 client nodes under a  label skewed scenario created for analysing the effectiveness of the Federated approach of model building. The results after the third communication round of global update to the clients are presented in Figure \ref{fig:CLU1}. Performance using a centralised model was also captured in order to compare the performance on the designated test images for both centralised and Federated Learning approaches. The cross entropy loss for 50 epochs can be observed for the 3 clients and for the centralised approach. The performance was found to be more than 95\% in all cases for 10-fold cross validation. Similarly, the performance of CityScape Dataset is presented in  Figure \ref{fig:CSP1}. A sample size of 118 images were used in a label skewed scenario distributed across 3 clients. 
For the FedUKD experiment, there were 10 communication rounds each with two epochs of training. The graph of the results represents cumulative epochs (Figure \ref{fig:FedUKDCS}). Communication and Federated Averaging happens after every two epochs. For best results the temperature was set as five and the learning rate alpha to 0.3. Both CityScape and Chennai Land Use datasets were pre-processsed to be non-IID as described in Scetion 3. 
On the CityScape dataset, teacher was trained on a large dataset (2975 training samples, 500 validation sample) for just two epochs. The accuracy was 72\% for training and 66\% for validation. The student model was trained on the smaller Non-IID dataset consisting of only 98 images. Max training accuracy after Federated Learning was 71\% in client 1 while max validation accuracy was 55\%. The Teacher model had 17 million parameters which represented 69 MBs of data to be transferred over the network. The student model on the other hand only 1 million parameters which only took 4 MBs of space. We were able to compress the data by almost 17 times.
The results were even better with the Chennai Land Use dataset (Figure \ref{fig:FedUKDCLU}). The accuracy was 97\% and we were able to shrink the the 17 million model parameters to 270 thousand which only represented 1 MB of data to be transferred, a optimization of over 62 times in terms of the number of parameters and 69 times in terms of space. Parameter optimization is the reduction in the number of parameters while space optimization is the reduction in the amount of memory taken by those parameters for example we reduced the space requirement from 69 MBs to 1 MB. 

\addtolength{\tabcolsep}{-3pt}
 \begin{table}[htbp]
\caption{Model Performance: Summary of Results}
\begin{center}
\scriptsize
\begin{tabular}{|c|c|c|c|c|c|}
\hline
Model & Federated & Dataset & Accuracy & Parameter  & Space \\
 &  &  & &  Optimization &  Optimization \\
\hline
UNet & No & CLU & 95\% & 1x & 1x\\
\hline
UNet & No & CSP & 72\% & 1x & 1x\\
\hline
FedUNet & Yes & CLU & 95\% & 1x & 1x\\
\hline
FedUNet & Yes & CSP & 78\% & 1x & 1x\\
\hline
FedUKD & Yes & CLU & 97\% & 62x & 69x\\
\hline
FedUKD & Yes & CSP & 71\% & 17x & 17.25x \\
\hline
\end{tabular}
\label{tab1}
\end{center}
\end{table}

\normalsize
\par We summarize the benefits of using knowledge distillation with our federated UNet below:
\begin{enumerate}
    \item Clients are located far away in the case of urban planning, since they can span entire cities. This makes communication slow and expensive especially for large models. Our FedUKD model is able to shrink down the data needed to be transferred by over 69 times, making communication quick and effective.
    \item The student model can achieve an accuracy comparable to the Teacher model, as demonstrated by our experiment on the Chennai Land Use dataset.
    \item Model might converge faster (in fewer epochs and thus less time) because the number of parameters to train is much smaller.
    \item In case a big dataset is not available and clients don’t want to share data, the Teacher can be trained on bigger public dataset and then used to train student on smaller non-IID dataset available to the clients.
    \item Our approach can also prevent certain clients from overfitting, for example client 2 had a very small number of samples and hence was overfitting around epoch 6 but after the federated averaging step the client 2's loss was aligned with the loss from the other clients. 
\end{enumerate}

\textbf{Anonymized code link:}
\url{https://anonymous.4open.science/r/FedUKD}\\
\textbf{Anonymized link for Chennai Land Use (CLU) Dataset:}\\
\url{https://anonymous.4open.science/r/ChennaiLandUseDataset/}

\normalsize
%\newpage
\section{Conclusions}

Remote monitoring of climate change globally is the way forward to address the growing impact on the environment and our planet. The 196 countries who agreed to the Paris climate accord in 2015, pledged to keep the global average temperature rise to below 1.5°C and reduce greenhouse gas emissions. But how can these promises be measured and monitored for compliance both locally and globally? In this paper we propose a novel strategy using state-of-the-art Federated Learning approach to design and develop a Deep Learning model using Knowledge Distillation for Land Use classification from satellite and street view images. The advantage of using a Federated approach for this application, is primarily to save communication cost by carrying out localized training and avoiding transfer of data to a centralized location. Federated Learning efficiently handles the inherently non-IID data across clients during the distributed model building process.  We created our own dataset for the study and also used bench-marked dataset to evaluate the performance of our model. Since impact on environment and climate change can be inferred from Land Use, a robust Land Use monitoring systems can help to bring in accountability in climate change management across the globe. The proposed model with more than 95\% accuracy can be integrated into computer vision based systems for monitoring the impact on the environment globally.
% \begin{figure}
%      \centering
%      \begin{subfigure}[b]{0.20\textwidth}
%       \includegraphics[width=\linewidth]{FedUKDCS.png}
%       \caption{CityScape Dataset}
%       \label{fig:FedUKDCS}
%       \end{subfigure}
%      \begin{subfigure}[b]{0.20\textwidth}
%       \includegraphics[width=\linewidth]{FedUKDChennai.png}
%       \caption{Chennai Land Use Dataset}
%       \label{fig:FedUKDCLU}
%       \end{subfigure}
%      \caption{FederatedUKD - Model Performance}
% \end{figure}

\normalsize

\backmatter

%%===================================================%%
%% For presentation purpose, we have included        %%
%% \bigskip command. please ignore this.             %%
%%===================================================%%
%\bigskip
%\begin{flushleft}%
%Editorial Policies for:

%\bigskip\noindent
%Springer journals and proceedings: %\url{https://www.springer.com/gp/editorial-policies}

%\bigskip\noindent
%Nature Portfolio journals: \url{https://www.nature.com/nature-research/editorial-policies}

%\bigskip\noindent
%\textit{Scientific Reports}: \url{https://www.nature.com/srep/journal-policies/editorial-policies}

%\bigskip\noindent
%BMC journals: \url{https://www.biomedcentral.com/getpublished/editorial-policies}
%\end{flushleft}

%%===========================================================================================%%
%% If you are submitting to one of the Nature Portfolio journals, using the eJP submission   %%
%% system, please include the references within the manuscript file itself. You may do this  %%
%% by copying the reference list from your .bbl file, paste it into the main manuscript .tex %%
%% file, and delete the associated \verb+\bibliography+ commands.                            %%
%%===========================================================================================%%

\bibliography{urban-ref}% common bib file
%% if required, the content of .bbl file can be included here once bbl is generated
%%\input sn-article.bbl

%% Default %%
%%\input sn-sample-bib.tex%

\end{document}